\documentclass[letterpaper]{article}
\usepackage{kevinpaper}
\usepackage{graphicx}
\usepackage{graphics}
\usepackage{subfigure}
\usepackage{kevin}
\usepackage{color}
\usepackage{setspace}
\usepackage{wrapfig}
\usepackage{fancybox}
\DeclareGraphicsExtensions{.jpg,.pdf,.tif,.png,.tiff}
\title{Using Covariance Matrices as Feature Descriptors for Vehicle Detection from a Fixed Camera}
\author{Kevin Mader \and Gil Reese\\
\emph{Department of Electrical and Computer Engineering, Boston University}\\
December 6th, 2006}
\pubnote{\em Cars and Trucks - Kevin Mader and Gil Reese}
\begin{document}
\titleboxpage
\newpage
\oddsidemargin -0.5in           
\textwidth 7.45in              

\topmargin 0in                 

\headheight 12pt               
\headsep 24pt                  

\textheight 8.4in
\footskip 0.4in
\begin{spacing}{.9}
\maketitle
\begin{abstract}
\indent

A method is developed to distinguish between cars and trucks present in a video feed of a highway. The method builds upon previously done work using covariance matrices as an accurate descriptor for regions. Background subtraction and other similar proven image processing techniques are used to identify the regions where the vehicles are most likely to be, and a distance metric comparing the vehicle inside the region to a fixed library of vehicles is used to determine the class of vehicle.
\end{abstract}
\end{spacing}
\section{Introduction}
\indent There are many potential uses for object identification ranging from development of automatic key-wording in a photo library to determining the weight load on a bridge or specific road or possibly estimating air pollution by the different types of vehicles present. The specific goal for this paper is to develop a method to distinguish and count the number of cars and trucks on the road at a given time.
\\ \indent
The problem of identifying vehicles has been mostly related to tracking uses, but several have proposed interesting approaches to the problem. Methods such as deformable template matching and template differencing \cite{mvdt} have been used for problems that necessitated real-time algorithms. Slower methods using histograms in the wavelet domain were able to detect objects from a variety of viewing angles \cite{stat3d}.
\\ \indent
The method we used was derived largely from a paper about covariance matrices as a distance metric \cite{covpaper}. However since the specific problem we had of identifying vehicles in a feed of images, we decided to utilize the information that could be obtained for comparing temporally separated frames. So instead of using region growing methods to identify the vehicles we used background subtraction and simple image segmentation methods to identify the regions where vehicles were located. Then we used covariance matrices on these regions to determine what kind of vehicle if any was present in the bounding box. \\
\indent One of the decisions that must be made before beginning to do any kind of analysis is to determine what are the groups vehicles are being classified into. Our selection was to define smaller vehicles as cars and larger vehicles as trucks with the cutoff being around the size of a midsize SUV such as a Chevrolet Tahoe. The decision was somewhat arbitrary, but the general reasoning behind it was caused by initially the small number of trucks present in the data set we chose which would be prohibitive of doing extensive testing. Secondly though was the simplicity of the separation of car and semi-truck. The two objects are vastly different in size and demonstrating a complex method using covariance matrices could separate these two classes of vehicles would be minimally informative since a simple pixel counting method could perform the task. Thirdly choosing groups that were closer together could allow us to investigate what kind of shape information in the covariance matrices allows the algorithm to make a decision to classify a vehicle as either a car or truck.

\section{Material and Methods}
\subsubsection{Image Acquisition} 
The images used for this project were obtained from three separate web cams (AXIS 207W Network Camera) placed in different rooms and at different viewing angles overlooking the Massachusetts Turnpike (I-90) a 6 lane interstate with a variety of vehicle traffic. The images were taken during a relatively low traffic period to minimize ghosting effects and overlapping vehicles and in most occasions the ambient light present was low enough to not cause excess glare or amplify the dirt on the windows. The camera produced jpeg images with a resolution of 320$\times$240 at a frame rate of around 10 frames per second. The different angles of the cameras provided a slightly different view of the scenery especially of a light pole that partially obstructed the view of the cars. In one camera the light pole was almost unnoticeable in the other two it blocked a part of the road so cars going under the pole were partially obstructed.
\subsubsection{Image Preprocessing}
A test image from the camera is rotated and then cropped from manual user input in order to establish the road and setup the cars to be orthogonal to the viewing window so a bounding box is an accurate way to express the boundaries of the car. If the car was moving in a diagonal fashion the bounding box would contain a significantly larger amount of empty space which would 'water-down' many of the calculations later possibly enough to make distinction between a car and a truck impossible.
\subsubsection{Background Subtraction} 
In order to remove the information that does not pertain to the objects and vehicles in the image the average image from the data set is subtracted. Given a color image set $H$ of dimension $W \times H \times \emph{\# of Color Channels} \times Images$ we calculate the background for the data set by the following formula
\begin{equation}
B(x,y,c)=\frac{1}{N} \sum_{n=1}^{N} H(x,y,c,n)
\end{equation}
In order to calculate a given image to be used for in the processing the background image is subtracted and the imaged is converted to grayscale by averaging each of the color channels
\begin{equation}
 H^\prime_k(x,y) =\frac{1}{3}\sum_{c=1}^{3} \left| H(x,y,c,k)- B(x,y,c)\right|
\end{equation}
\subsubsection{Image Cleaning} 
The process of identifying objects and vehicles from the image $I_k(x,y)$ is a very difficult one, but is essential for determining a car or a truck. The first step is to create a cleaned up version of the image. This is accomplished by first using a median filter with a $[5\times5]$ neighborhood.
\begin{equation}
H^{\prime\prime}_k(x,y)=\mathbf{MedianFilter}(H^\prime_k(x,y))
\end{equation}
The next step is to remove all the low valued pixels by shifting the image down by 20 and removing all the negative numbers.
\begin{eqnarray}
I_k(x,y)=\left\{\begin{array}{cl}
	0, & H^{\prime\prime}_k(x,y)<10 \\
	H^{\prime\prime}_k(x,y)-10, & H^{\prime\prime}_k(x,y)\geq 10
	   \end{array}\right. 
\end{eqnarray}
\subsubsection{Binary Image}
A binary image is important for segmentation process since a grayscale image would contain too much information to use standard \texttt{MATLAB} methods. A new image is created from the image and an amplified version of the edges in the image.  The $\mathbf{Edges}$ term is binary image created from doing a canny edge detection on the image. 
\begin{equation}
I_k^\prime(x,y)=I_k(x,y)+10\times\mathbf{Edges}\{I_k(x,y)\}
\end{equation}
The image is then converted to a binary image using intensity cutoff
\begin{eqnarray}
I^{\prime\prime}_k(x,y)=\left\{\begin{array}{cl}
	0, & I_k^\prime(x,y)<\mu \\
	1, & I_k^\prime(x,y)\geq \mu
	   \end{array}\right. \\
\mu=\frac{1}{\text{N}\times \text{W} \times \text{H}} \sum_{k^\prime=1}^{\text{N}}\sum_{x^\prime=1}^{\text{W}}\sum_{y^\prime=1}^{\text{H}} I_{k^\prime}^\prime(x^\prime,y^\prime)
\end{eqnarray}
$\mu$ is the average value over all images in the data set to prevent images absent of cars to have the noise amplified disproportionately.
\subsubsection{Segmentation}
From the binary image separate objects are detected by finding all the pixels that formed contiguous groups of more than 60 pixels each of these groups was labeled as a region $i$ out of $L$, size $\text{W}_i$ by $\text{H}_i$, starting at point $(x_i,y_i)$ in image $I^{\prime\prime}_k(x,y) \rightarrow R_{i,k}(x,y)=I^{\prime\prime}_k(x_i+x,y_i+y) \forall [(x,y)\leqslant(\text{W}_i,\text{H}_i)]$ For each of these regions a mean (a mean in a binary image represents the percentage of pixels turned on) $f_i$was calculated as follows.
\begin{eqnarray}
f_i=\frac{1}{\text{W}_i \times \text{H}_i} \sum_{x^\prime=1}^{\text{W}_i}\sum_{y^\prime=1}^{\text{H}_i} R_{i,k}(x^\prime,y^\prime)
\end{eqnarray}
Since $I^{\prime\prime}_k(x,y)$ is a binary image and $R_{i,k}(x,y) \in I^{\prime\prime}_k(x,y)$ then $0 \leq f_i\leq1$. 
\begin{itemize}
\item If $f_i \leq 0.45$ and the value of $\text{W}_i\times \text{H}_i \geq 1400 px^2$ the object is assumed to contain multiple cars that are out of alignment and therefor have quite a bit of empty space resulting in a low $f_i$ value. In order to seperate these two objects a 'k-means algorithm' \cite{kms} is used to search for two groups of pixels inside the region $R_{i,k}$ from this two new regions are created.
\item If  $f_i \geq 0.80$ and the value of $\text{W}_i\times \text{H}_i \leq 340 px^2$ the object is assumed to contain a portion of a vehicle (windshield, hood, etc.). The assumption is then that there are nearby segments containing the other parts of this same car so a "k-means algorithm" \cite{kms} is used to search for $L-1$ groups.
\item If $\text{W}_i\times \text{H}_i \leq 200 px^2$ and none of the other conditions are true the section is deleted
\end{itemize}
 Each of these groups was taken and a bounding box was determined. If the number of  and the samples from $I_k(x,y)$ inside the box are used to calculate the covariance matrix.
\subsubsection{Feature Vectors}
The feature vector used in the Region Covariance paper \cite{covpaper} requires revision since our task requires correctly identifying cars from trucks. In that paper they were trying to identify specific objects that were moving around not classes of objects. So color information is going to be ignored as trucks are not significantly differently colored than cars. The most successful feature vectors used on our data set have been. 
\begin{center}
\begin{eqnarray}
\mathbf{F}_{x,y}=[
\begin{array}{cccccccc} 
x & y & \frac{\partial I(x,y) }{\partial x} & \frac{\partial I(x,y) }{\partial y} & \nabla^2 I(x,y)
\end{array}]\\
\mathbf{F}_{x,y}=[
\begin{array}{cccccccc} 
r^2 & \frac{\partial I(x,y) }{\partial x} & \frac{\partial I(x,y) }{\partial y} & \nabla^2 I(x,y)
\end{array}]\\
\mathbf{F}_{x,y}=[
\begin{array}{cccccccc} 
x & y & \nabla^2 I(x,y) & \mathbf{Edges}(I(x,y))
\end{array}]
\label{fvec}
\end{eqnarray}
\end{center}
$r=\sqrt{(x-x_0)^2+(y-y_0)^2}$ represents the distance the given pixel is away from the center of the region $(x_0,y_0)=(x_i+\frac{\text{W}_i}{2},y_i+\frac{\text{H}_i}{2})$. The use of R over x and y was investigated since using a rotation invariant variable would allow the car to be going the opposite direction or changing lanes and still produce a similar feature vector. The derivatives in the x and y directions are approximated by convolution with the 3$\times$3 matrix Sobel in x and y
\begin{eqnarray}
\frac{\partial}{\partial x}\approx\left[\begin{array}{ccc}
-1 & 0 & 1 \\
-2 & 0 & 2 \\
-1 & 0 & 1 \\
\end{array}\right] & \frac{\partial}{\partial y} \approx\left[\begin{array}{ccc}
1 & 2 & 1 \\
0 & 0 & 0 \\
-1 & -2 & -1 \\
\end{array}\right]
\end{eqnarray}
The laplacian is approximated by a convolution using the 3$\times$3 Laplacian matrix.
\begin{eqnarray}
\nabla^2\approx\left[\begin{array}{ccc}
-1 & -1 & -1 \\
-1 & 8 & -1 \\
-1 & -1 & -1 \\
\end{array}\right] 
\end{eqnarray}
\subsubsection{Covariance Matrices}
The covariance matrix $C_{k}$ for a given region $R_k(x,y)$ was calculated by creating a feature vector $\mathbf{F}_i$ for each point $(x,y) \rightarrow i=x+(y-1)*\text{W}_i$ in the region resulting in $N$ total points. $\mu(u)$ represents the average value of the feature value $u$ for all the pixels in the region.
\begin{eqnarray}
\mathbf{C}_{k}(u,v)=\frac{1}{\text{N}} \sum_{i=1}^{\text{N}} (\mathbf{F}_i(u)-\mu(u) )(\mathbf{F}_i(v)-\mu(v))
\end{eqnarray}
The distance between two covariance matrices is calculated using a previously developed method \cite{covdist}.
\begin{eqnarray}
\rho(\mathbf{C}_1,\mathbf{C}_2)=\sqrt{\sum_{i=1}^{n} \ln^2\lambda_i(\mathbf{C}_1,\mathbf{C}_2)}
\label{distn}
\end{eqnarray}
Where $\lambda_i(\mathbf{C}_1,\mathbf{C}_2)$ repesents the generalized eigenvalues between $\mathbf{C}_1$ and $\mathbf{C}_1$. While this metric is likely non-optimal, it satisfies the requirements for a distance metric and is easily and quickly calculated using \texttt{MATLAB}
\subsubsection{Ontology Creation}
One image set was selected as containing a large amount of diversity of vehicles and overlaps to develop the library of covariance matrices divided into 4 categories \{\emph{Car}, \emph{Truck}, \emph{Multiple}, \emph{Junk}\}. Multiple was created to represent when two vehicles were overlapping enough to be identified by the segmentation algorithm as the same vehicle. Junk was created when the algorithm would identify just a portion of a vehicle or when the vehicle is partially out of frame. The method then determined what class of object a given region $R_i$ contained by calculating the distance between $C_i$ and every matrix in the library. The minimum distance was then taken and the region would be identified as the same class as the object it was closest too.

\section{Results}
\indent Some of the results produced by the algorithm are listed in quantitative from in Table \ref{senz}. The only feature vector used to produce these specific results was Equation [\ref{fvec}]. The others were used initially but this one appeared to work the best. The sensitivity and specificity terms are defined in the appendix under Equations [\ref{sense},\ref{spec}] respectively.

\begin{table}
\caption{Quantitative Results}\label{senz}
\begin{tabular}{||c|c|c|c||} 
\hline
\small{Length} & \small{\# Cars} & \small{Sensitivity} & \small{Specificity}  \\
\hline \hline
\small{15 frames} & 16 & 93.75\% & 0\%\\
\hline
\hline
\hline
\small{Length} & \small{\# Trucks} & \small{Sensitivity} & \small{Specificity} \\
\hline \hline
\small{15 frames} & 2 & 100\% & 100\%\\
\hline
\hline
\end{tabular}
\end{table}

The table is from a fairly simple scene typical in low traffic conditions where there is minimal overlap of the vehicles, and for this scene it works very well only making a mistake on one frame and the mistake was in the segmentation (Figure \ref{badsplit}) the algorithm's only mistake was to label the two partial car segments both as cars instead of junk. The specificity number could probably be greatly increased by an improved segmentation.

\begin{figure}[p]
\centering
\subfigure[Good multiple car detection]{
\includegraphics[width=.55\textwidth]{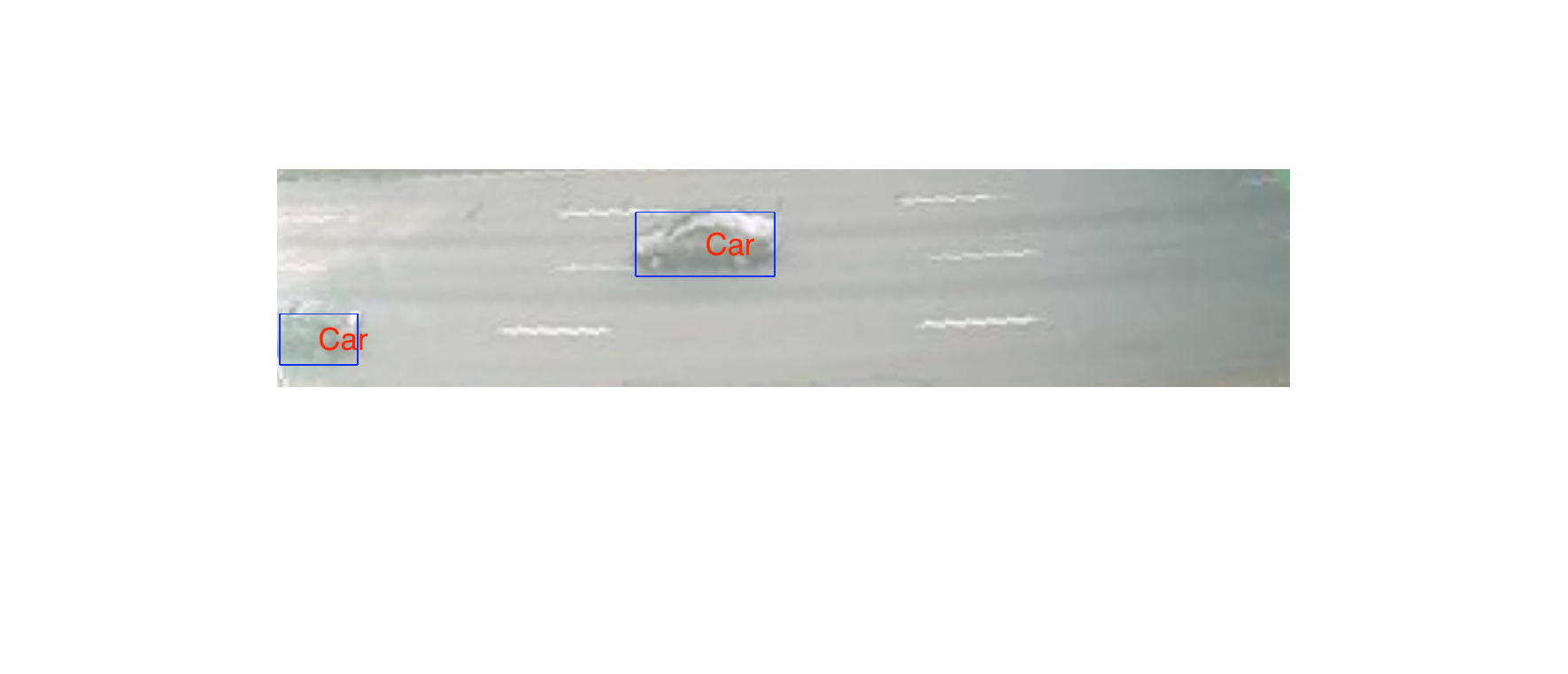}
\label{c1}
}
\subfigure[Good truck detection]{
\includegraphics[width=.55\textwidth]{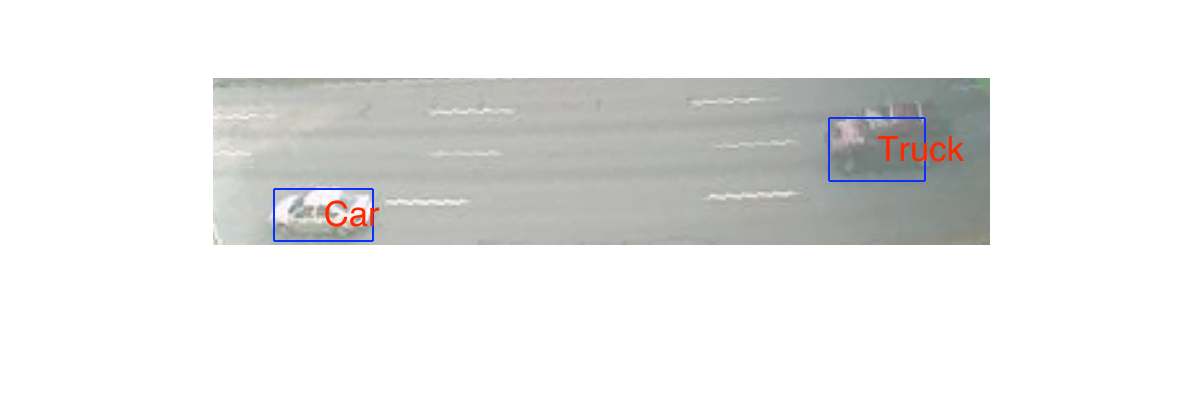}
\label{t1}
}
\subfigure[Overlapping Vehicle detection]{
\includegraphics[width=.55\textwidth]{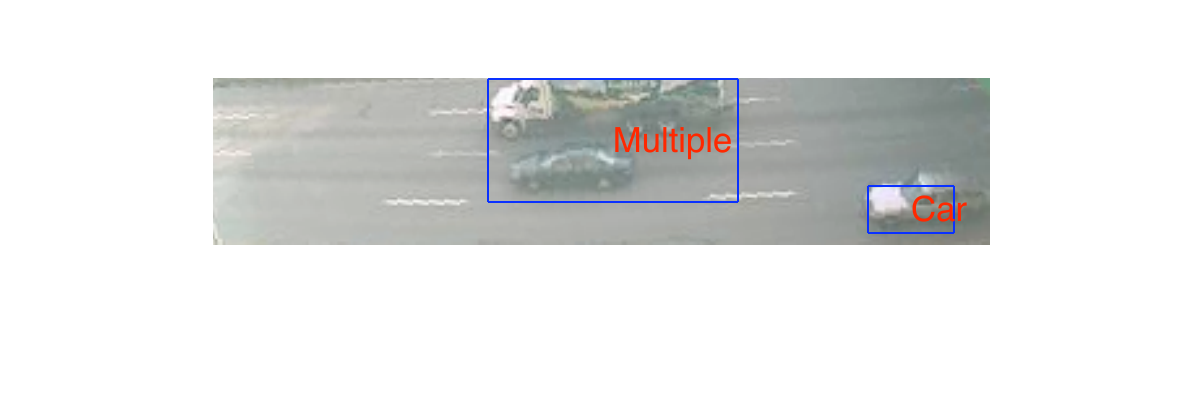}
\label{m2}
}
\subfigure[Good multiple object detection] {
\includegraphics[width=.55\textwidth]{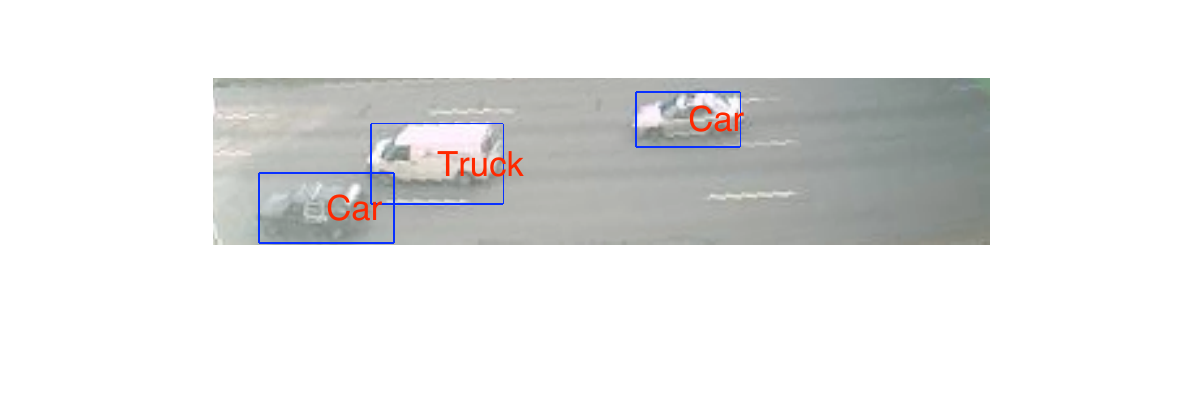}
\label{m1}
}
\caption{Normal Images}
\label{norm}
\end{figure}

\begin{figure}[p]
\centering
\subfigure[Good multiple car and junk detection]{
\includegraphics[width=.35\textwidth]{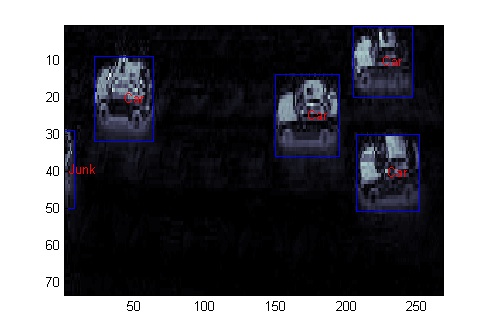}
\label{bkc1}
}
\subfigure[Good truck detection]{
\includegraphics[width=.35\textwidth]{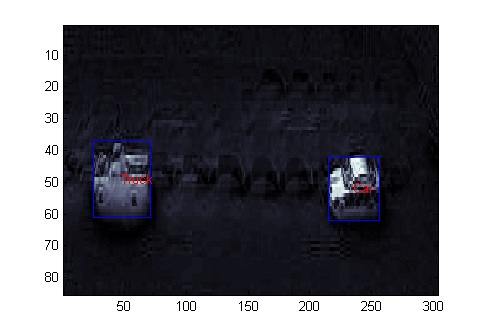}
\label{bkt1}
}
\subfigure[Good partial truck detection]{
\includegraphics[width=.35\textwidth]{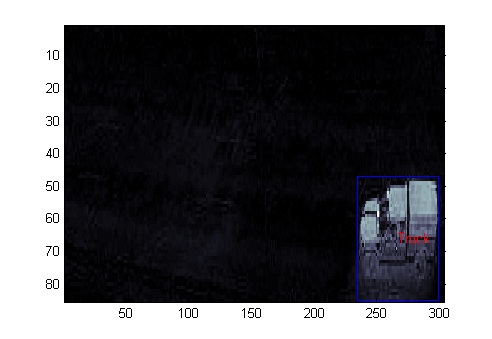}
\label{bkm2}
}
\subfigure[Good multiple object detection] {
\includegraphics[width=.35\textwidth]{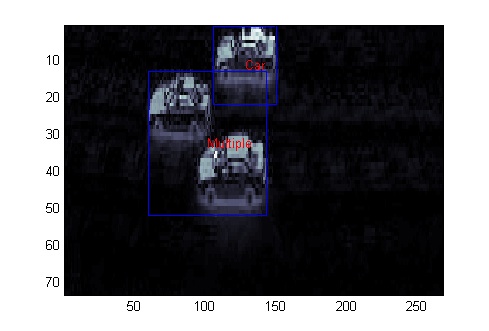}
\label{bkm1}
}
\caption{Background Subtracted Images}
\label{bk}
\end{figure}

Figures [\ref{norm},\ref{bk}] demonstrate the algorithm's ability to detect all the objects in the scene and properly identifying the cars. The pictures in Figure [\ref{bk}] show the background subtracted images with the bounding box and indentification. The ghosting sort of effects that appear as the image set gets small appears strongly in images [\ref{bkt1},\ref{bkm1}]. In this image set the effect was weak enough to not cause false identification of objects, but the ghosting effects can cause quite a few issues as discussed later.

\section{Discussion}
\begin{figure}[hbp]
\centering
\subfigure[Car detection with pole obstruction]{
\includegraphics[width=90mm]{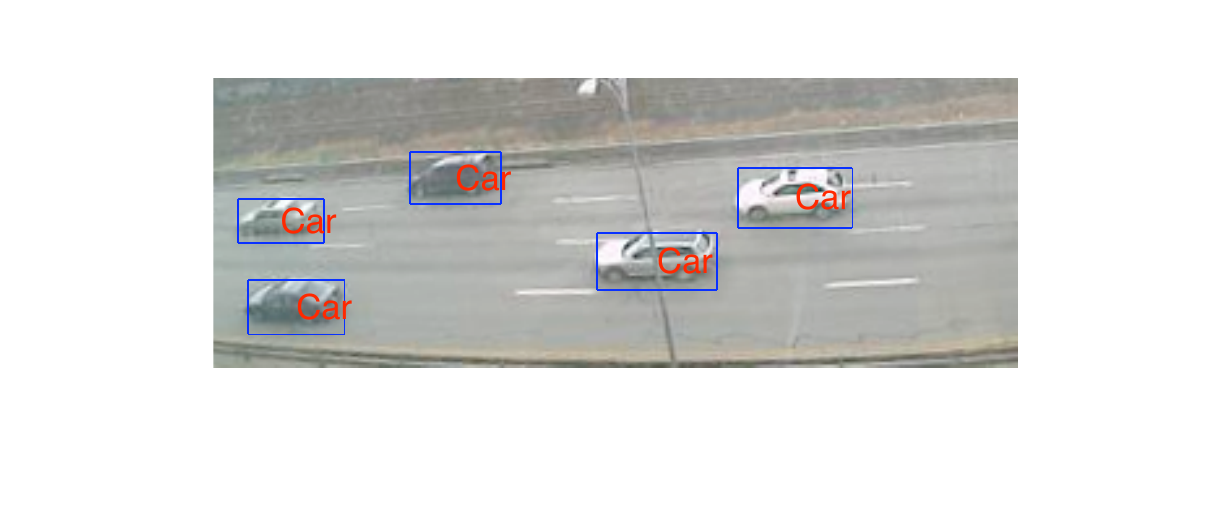}
\label{cc1}
}
\subfigure[3 overlapping cars detection and junk detection]{
\includegraphics[width=90mm]{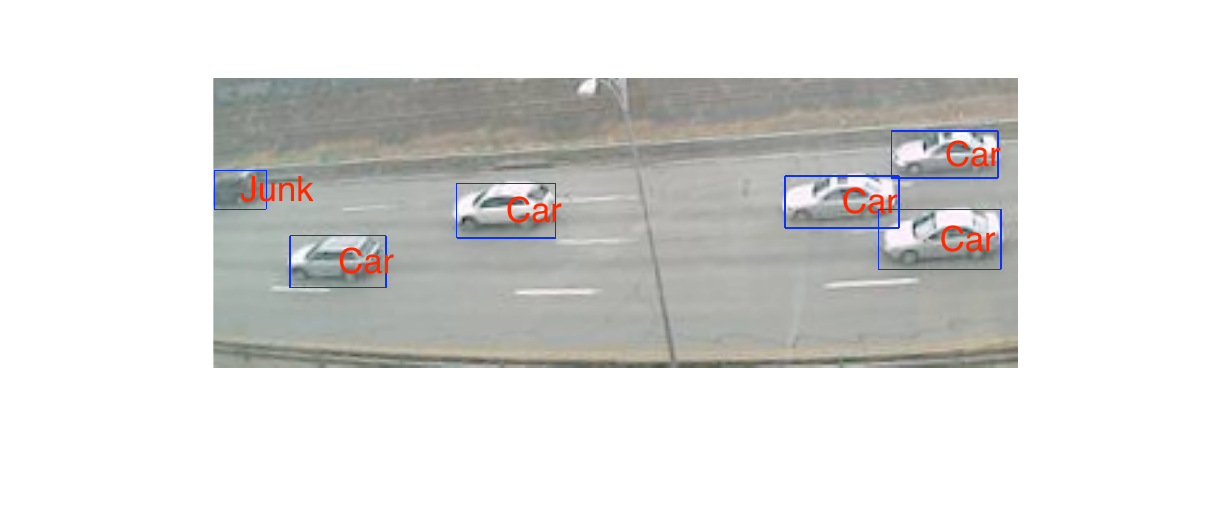}
\label{cc2}
}
\subfigure[Truck detection]{
\includegraphics[width=90mm]{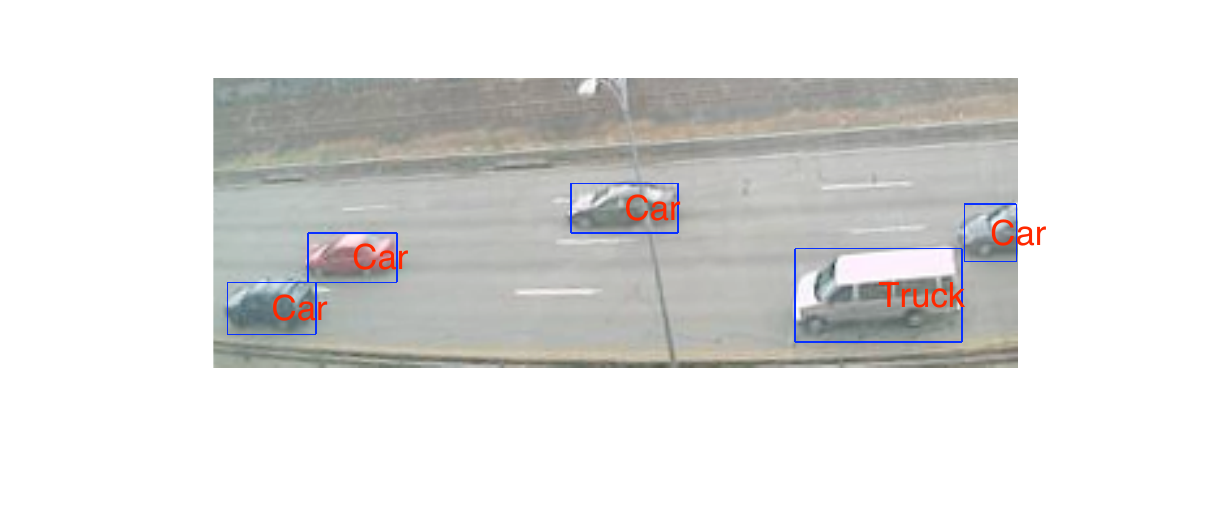}
\label{cct}
}
\subfigure[Truck detection with pole obstruction] {
\includegraphics[width=90mm]{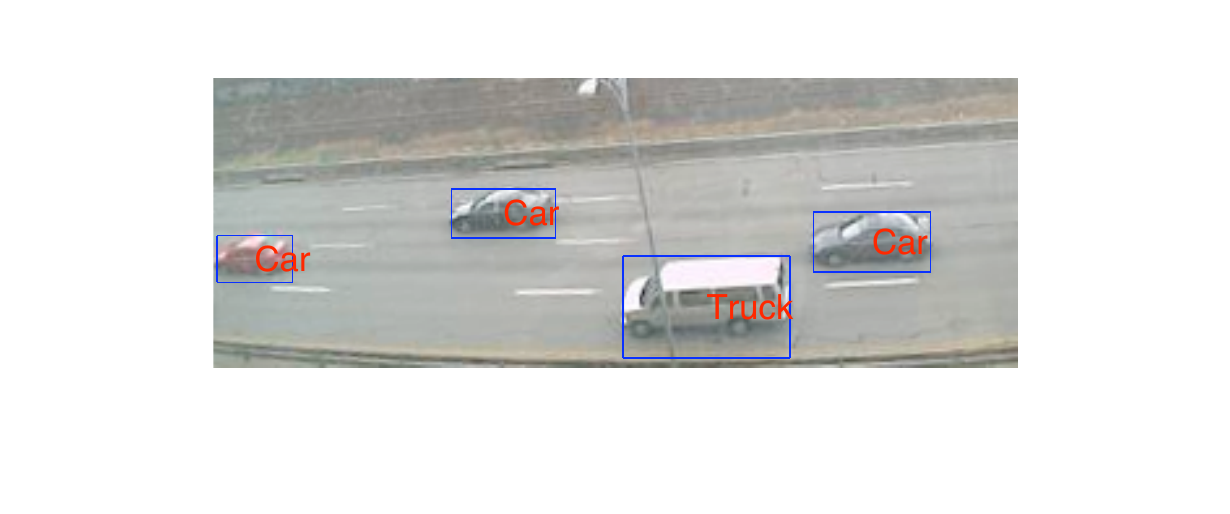}
\label{cct2}
}
\caption{Difficult Images}
\label{cc}
\end{figure}

\indent The algorithm worked suprisingly well on most of the images even difficult images as demonstrated in Figure \ref{cc}. The program seemed to work very well regardless of how the camera angles and lighting changed from image set to image set. Figure \ref{cc} shows the correct identification of all vehicles in the picture using an ontology that was created using a camera looking the other direction. 
The algorithm success was almost entirely limited by the success of the segmentation algorithm on the image.
\subsubsection{Shortcomings}
\begin{figure}[t]
\includegraphics[width=.4\textwidth]{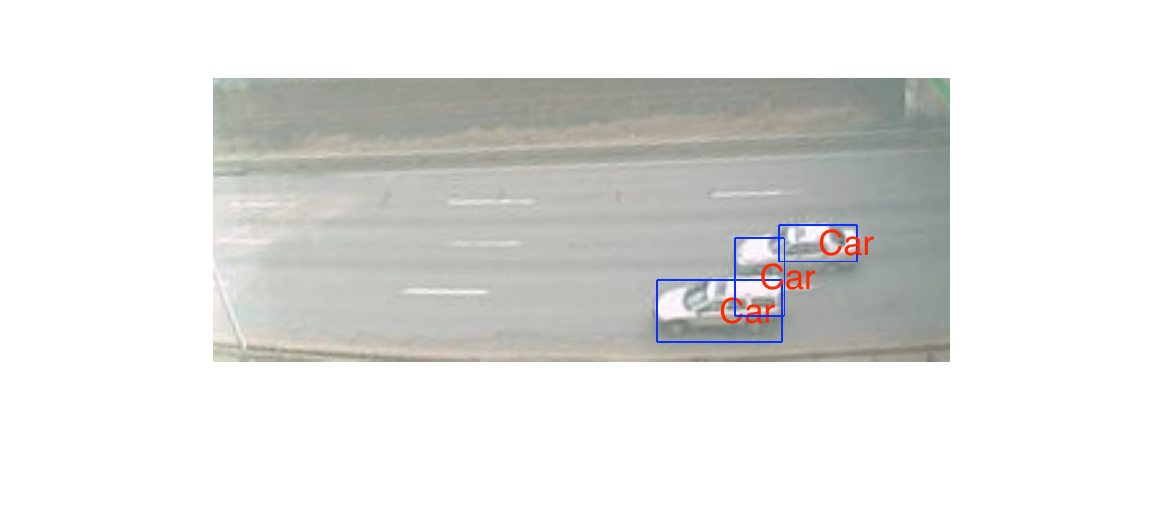}
\caption{The two vehicles here were split into 3 vehicles}
\label{badsplit}
\end{figure}

\begin{figure}[h]
\includegraphics[width=.4\textwidth]{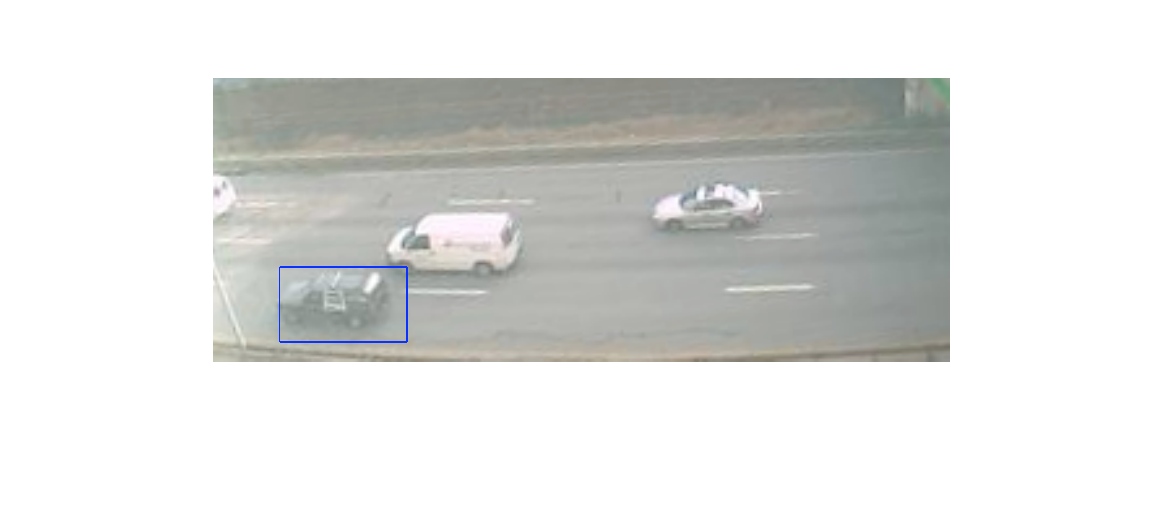}
\caption{The vehicle equidistant (using Equation [\ref{distn}]) between a car and a truck (the 'eigenvehicle')}
\label{eqidist}
\end{figure}

The segmentation algorithm was able to manage most of the images very well; however, several of the images managed to produce erroroneous regions that either contained multiple cars (Figure \ref{bkm1}) or regions that contained portions of cars (Figure \ref{badsplit}).
\indent One of the largest potential problems with this method is the distances between objects within the library. Small distances between them would mean the difference in distances between a given region and an object in the car class and an object in the truck class could be quite small. With these differences being quite small the proper identification could be thrown off my small noise artifacts in the image. Although there were some cases such as Figure [\ref{eqidist}, \ref{stationwagon}] where the distance metric was not distinct enough to separate the vehicles in most cases it seemed to work well.\\
One of the most interesting failures of the algorithm that was actually found because of a bug in the program was an object that was eqi-distant from both the car and the truck using the metric (Eq \ref{distn}). The picture of this object is shown in (Figure \ref{eqidist}).
Lines on the road show up on the background subtracted images as lines on the vehicle. This causes possible extra edges to appear inside the car where there is nothing. This effect was noticeable in several of the images, but it did not seem to pose a problem when classifying the type of image. However it is fairly easy to imagine a configuration where this could pose a real problem.
\subsubsection{What is a truck?}
\begin{figure}[h]
\includegraphics[width=.4\textwidth]{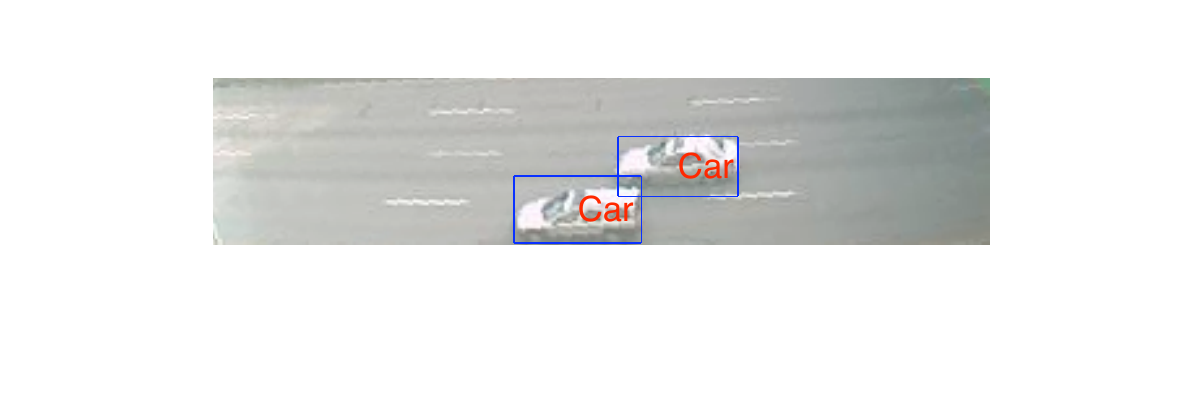}
\caption{Station wagons were often inconsistently classified from frame to frame}
\label{stationwagon}
\end{figure}

For the purposes of our project a truck was really just defined as a vehicle that was noticeably larger than a sedan. So a Jeep or small pickup would not be considered a truck, but a Hummer, passenger van, and large pickup would be. Clearly a large vehicle or semi-truck would always be considered a truck, and a small sedan or a 5 passenger SUV would be considered a car. This method worked well on the vehicles which fell on the extremes of being a car or a truck but several vehicles in between were poorly identified by the algorithm. The vehicle would switch between car and truck since the distance difference between the two was quite small. A great example of this is the station wagon (Figure \ref{stationwagon}). Furthermore many car manufacturers today develop trucks to look like cars and cars to look like trucks so one would expect that these vehicles might be classified incorrectly which again reiterates the importance of having a solid definition of vehicle types for classification. 
\subsubsection{Assumptions and Limitations}
The first assumption is the camera is fixed so the rotation and cropping procedure needs to only be done once per camera and within an image set the only changes would be related to vehicle motion. Some assumptions being made about the size of the image set is that it is large enough that the average image is an accurate representation of the background, yet small enough in comparison to the time-scale of a day to be significantly altered by sunrise or sunset. A time sample of up to about 15 to 30 minutes is about the maximum time length for this invariance to be true and a sample of at least 50 images is required for the average image to be free of ghosting. Additionally the cars are assumed to be moving quickly only in view for several seconds at most. Conditions such as bumper to bumper traffic would give a poor average image and a significant amount of vehicle overlap.
\section{Future Work}
\indent A deeper investigation needs to be done on what really defines a car or truck from the covariance statistics. This would allow us to understand exactly which variables play the largest roles in the determination of a vehicle, furthermore this might allow for the development of a better suited metric for measuring distance instead of logs of eigenvalues it might be possible to compare several elements of the covariance matrix. 
\\ \indent For our project the algorithm was qualitatively tweaked and optimized. While this is much easier to do than developing quantitative methods, it is probably flawed. The algorithm needs to be tested more extensive and statistics about sensitivity and specificity for car and truck detection need to be determined for this method to be useful in a real world setting. The algorithm also needs to be compared to other simpler algorithm's to determine its relative ability to properly identify cars. For example in many of the image sets it appeared that a measurement of the bounding box area would have been a closer match to the vehicle type than using the significantly more complicated and computationally intensive covariance matrix method. Also the problem is probably not properly defined. There are many more classes of vehicles than we described. There are mid-size, motorcycles, and all sorts of different SUV's that were lumped into either car or truck based on their relative size which probably weakened the algorithm's ability to discern between these middle class of vehicles.
\bibliographystyle{aaai}
\bibliography{projectpaper}
\onecolumn
\pubnote{\em \textbf{Appendix} - Cars and Trucks - Kevin Mader and Gil Reese \hfill A}
\setcounter{page}{1}
\section{Appendix}
\subsection{Sensitivity and Specificity}
Although these are generally used for medical statistics they have been used as a good measure for the tests especially in reference to proper junk identification when segmentation produces faulty regions. For these test the gold standard is a human user looking at each frame and identifying the number of cars and trucks they see, and then looking and how the program identified everything to determine the number of junk regions labeled as car or truck. A junk region would be all three regions in the splitting picture since none of them represent accurately a car (Figure \ref{badsplit}). So the sensitivity test tests the both the segmentation and accuracy of the minimum distance finding/covariance matrix ontology. The specificity test just shows the accuracy of the minimum distance finding/covariance matrix ontology because a statistic representing how often the algorithm correctly identified the absence of a car would be too saturated because the test does work in most cases. The specificity is just how well the program labeled the regions that were found as junk. For example a region that just contained a roof would be junk or contained two portions of different cars should be junk.

\begin{eqnarray}
\label{sense}
\texttt{Sensitivity}=\tiny{\frac{\emph{Correctly Identified Cars/Trucks}}{\emph{Total Number of Cars/Trucks}}}\\
\label{spec}
\texttt{Specificity}=\tiny{\frac{\emph{Correctly Identified Junk}}{\emph{Junk Labeled as Cars/Trucks}+\emph{Correct Junk}}}
\end{eqnarray}

\subsection{Background Subtraction/Cleaining Code}
\texttt{\hspace{1mm}\textcolor{blue}{function} [jk,rk]=loaddata() \\ 
\hspace{1mm}d=menu(\textcolor{red}{'Select File Loading Method'},\textcolor{red}{'Folder'},\textcolor{red}{'File(s)'}); \\ 
\hspace{1mm}\textcolor{blue}{if} d==1 \\ 
\hspace{1mm}\indent path=[uigetdir \textcolor{red}{'$\backslash$'}]; \\ 
\hspace{1mm}\indent files=dir([path \textcolor{red}{'*.jpg'}]); \\ 
\hspace{1mm}\indent filen={files.name}; \\ 
\hspace{1mm}\textcolor{blue}{else} \\ 
\hspace{1mm}\indent [filen,path]=uigetfile({\textcolor{red}{'*.jpg;*.JPG'},\textcolor{red}{'JPEG Image'}},\textcolor{red}{'Multiselect'},\textcolor{red}{'on'}); \\ 
\hspace{1mm}\indent \textcolor{blue}{if} ischar(filen) \\ 
\hspace{1mm}\indent \indent \indent filen={filen}; \\ 
\hspace{1mm}\indent \textcolor{blue}{end} \\ 
\hspace{1mm}\textcolor{blue}{end} \\ 
\hspace{1mm}test=imread([path filen{1}]); \\ 
\hspace{1mm}imshow(test) \\ 
\hspace{1mm}uiwait(msgbox(\textcolor{red}{'Click the bottom line of the image'})); \\ 
\hspace{1mm}[y,x]=ginput(2) \\ 
\hspace{1mm}thet=180/pi*atan(diff(x)/diff(y)); \\ 
\hspace{1mm}test=imrotate(test,thet,\textcolor{red}{'bicubic'}); \\ 
\hspace{1mm}imshow(test); \\ 
\hspace{1mm}uiwait(msgbox(\textcolor{red}{'Click the boundaries of the image'})); \\ 
\hspace{1mm}[y,x]=ginput(2) \\ 
\hspace{1mm}x=round(x); \\ 
\hspace{1mm}y=round(y); \\ 
\hspace{1mm}k=[]; \\ 
\hspace{1mm}\textcolor{blue}{for} jb=1:length(filen) \\ 
\hspace{1mm}\indent test=imread([path filen{jb}]); \\ 
\hspace{1mm}\indent test=imrotate(test,thet,\textcolor{red}{'bicubic'}); \\ 
\hspace{1mm}\indent k(:,:,:,jb)=double(test(min(x):max(x),min(y):max(y),:)); \\ 
\hspace{1mm}\textcolor{blue}{end} \\ 
\hspace{1mm}test=mean(k,4); \\ 
\hspace{1mm}jk=[]; \\ 
\hspace{1mm}rk=[]; \\ 
\hspace{1mm}\textcolor{blue}{for} jb=1:length(filen) \\ 
\hspace{1mm}\indent cImg=abs(k(:,:,:,jb)-test); \\ 
\hspace{1mm}	jk(:,:,jb)=mean(cImg,3); \textcolor{green}{\% convert to psuedo grayscale }\\ 
\hspace{1mm}\indent findcars(jk(:,:,jb)); \\ 
\hspace{1mm}\indent pause(.1); \\ 
\hspace{1mm}\textcolor{blue}{end} \\ 
\hspace{1mm}rk=uint8(k); \\ 
\hspace{1mm} \\ 
}
\subsection{Ground Truth/Ontology Creating Tool}
\texttt{\hspace{1mm}\textcolor{blue}{function} [imdb,obdb]=labelcars(imgs,realim,obdb,imdb) \\ 
\hspace{1mm}\textcolor{blue}{if} nargin$\leq$3 \\ 
\hspace{1mm}\indent \textcolor{green}{\%cardb }\\ 
\hspace{1mm}\indent obdb{1}.covs=[]; \\ 
\hspace{1mm}\indent imdb{1}=[]; \\ 
\hspace{1mm}\indent \textcolor{green}{\%truckdb }\\ 
\hspace{1mm}\indent obdb{2}.covs=[]; \\ 
\hspace{1mm}\indent imdb{2}=[]; \\ 
\hspace{1mm}\indent \textcolor{green}{\%multidb }\\ 
\hspace{1mm}\indent obdb{3}.covs=[]; \\ 
\hspace{1mm}\indent imdb{3}=[]; \\ 
\hspace{1mm}\indent \textcolor{green}{\%junkdb }\\ 
\hspace{1mm}\indent obdb{4}.covs=[]; \\ 
\hspace{1mm}\indent imdb{4}=[]; \\ 
\hspace{1mm}\textcolor{blue}{end} \\ 
\hspace{1mm} \\ 
\hspace{1mm}colormap(\textcolor{red}{'bone'}); \\ 
\hspace{1mm}\textcolor{blue}{for} k=1:size(imgs,3) \\ 
\hspace{1mm}\indent [xmin,xmax,ymin,ymax]=findcars(imgs(:,:,k),0); \\ 
\hspace{1mm}\indent fk=featureim(imgs(:,:,k)); \\ 
\hspace{1mm}\indent \textcolor{blue}{for} j=1:length(xmin) \\ 
\hspace{1mm}\indent \indent \indent imshow(realim(:,:,:,k)); \\ 
\hspace{1mm}\indent \indent \indent line([xmin(j) xmax(j)],[ymin(j) ymin(j)]); \\ 
\hspace{1mm}\indent \indent \indent line([xmin(j) xmax(j)],[ymax(j) ymax(j)]); \\ 
\hspace{1mm}\indent \indent \indent line([xmin(j) xmin(j)],[ymin(j) ymax(j)]); \\ 
\hspace{1mm}\indent \indent \indent line([xmax(j) xmax(j)],[ymin(j) ymax(j)]); \\ 
\hspace{1mm}\indent \indent \indent \textcolor{green}{\%[xmin(j),xmax(j),ymin(j),ymax(j)] }\\ 
\hspace{1mm}\indent \indent \indent \textcolor{green}{\%figure }\\ 
\hspace{1mm}\indent \indent \indent \textcolor{green}{\%imagesc(imgs(ymin(j):ymax(j),xmin(j):xmax(j),k)) }\\ 
\hspace{1mm} \\ 
\hspace{1mm} \\ 
\hspace{1mm}\indent \indent \indent fi=fk(ymin(j):ymax(j),xmin(j):xmax(j),:); \\ 
\hspace{1mm}\indent \indent \indent c=cov(reshape(fi,size(fi,1)*size(fi,2),size(fi,3))); \\ 
\hspace{1mm}\indent \indent \indent a=menu(\textcolor{red}{'Identify Object'},\textcolor{red}{'Ignore'},\textcolor{red}{'Car'},\textcolor{red}{'Truck'},\textcolor{red}{'Multiple'},\textcolor{red}{'Junk'})-1; \\ 
\hspace{1mm}\indent \indent \indent \textcolor{blue}{if} a$\ge$0 \\ 
\hspace{1mm}\indent \indent \indent \indent ing=imgs(ymin(j):ymax(j),xmin(j):xmax(j),k); \\ 
\hspace{1mm} \\ 
\hspace{1mm}\indent \indent \indent \indent \textcolor{blue}{if} isempty(obdb{a}.covs) \\ 
\hspace{1mm}\indent \indent \indent \indent \indent obdb{a}.covs(:,:,end)=c; \\ 
\hspace{1mm}\indent \indent \indent \indent \indent imdb{a}.imgs{1}.img=ing; \\ 
\hspace{1mm}\indent \indent \indent \indent \textcolor{blue}{else} \\ 
\hspace{1mm}\indent \indent \indent \indent \indent obdb{a}.covs(:,:,end+1)=c; \\ 
\hspace{1mm}\indent \indent \indent \indent \indent imdb{a}.imgs{end+1}.img=ing; \\ 
\hspace{1mm}\indent \indent \indent \indent \textcolor{blue}{end} \\ 
\hspace{1mm}\indent \indent \indent \textcolor{blue}{end} \\ 
\hspace{1mm}\indent \textcolor{blue}{end} \\ 
\hspace{1mm}\textcolor{blue}{end} \\ 
}
\subsection{Image Cleaning and Segmentation Code}
\texttt{\hspace{1mm}\textcolor{blue}{function} [xmin,xmax,ymin,ymax]=findcars(imgin,dp) \\ 
\hspace{1mm}\textcolor{blue}{if} nargin$<$2 \\ 
\hspace{1mm}\indent dp=1; \\ 
\hspace{1mm}\textcolor{blue}{end} \\ 
\hspace{1mm}imgin=medfilt2(imgin.*(imgin$>$10)-10,[5 5]); \\ 
\hspace{1mm}img=imgin+edge(imgin,\textcolor{red}{'canny'},.2)*10; \\ 
\hspace{1mm}bimg=bwareaopen(img$>$15,80); \\ 
\hspace{1mm}xmin=[]; \\ 
\hspace{1mm}xmax=[]; \\ 
\hspace{1mm}ymin=[]; \\ 
\hspace{1mm}ymax=[]; \\ 
\hspace{1mm}[bimg,n]=bwlabel(bimg); \\ 
\hspace{1mm}\textcolor{blue}{if} dp \\ 
\hspace{1mm}\indent imagesc(bimg); \\ 
\hspace{1mm}\textcolor{blue}{end} \\ 
\hspace{1mm}k=1; \\ 
\hspace{1mm}\textcolor{blue}{while} k$<$=n \\ 
\hspace{1mm}\indent tImg=(bimg==k); \\ 
\hspace{1mm}\indent xI=find(mean(tImg)$>$0); \\ 
\hspace{1mm}\indent yI=find(mean(tImg')$>$0); \\ 
\hspace{1mm}\indent ttImg=tImg(min(yI):max(yI),min(xI):max(xI)); \\ 
\hspace{1mm}\indent pn=mean(mean(ttImg)); \\ 
\hspace{1mm}\indent \textcolor{blue}{if} pn$<$.45 \& (size(ttImg,1)*size(ttImg,2))$>$350*4 \textcolor{green}{\% probably 2 vehicles, lots of empty space }\\ 
\hspace{1mm}\indent \indent \indent disp(\textcolor{red}{'KMean Split'});   \\ 
\hspace{1mm}\indent \indent \indent ibImg=ikmeans(tImg,2); \\ 
\hspace{1mm}\indent \indent \indent [x,y]=find(bimg==k); \\ 
\hspace{1mm}\indent \indent \indent bimg(x,y)=0; \textcolor{green}{\% delete the current one }\\ 
\hspace{1mm}\indent \indent \indent bimg=bimg+(ibImg==1)*k; \\ 
\hspace{1mm}\indent \indent \indent bimg=bimg+(ibImg==2)*(n+1); \\ 
\hspace{1mm}\indent \indent \indent imagesc(bimg) \\ 
\hspace{1mm}\indent \indent \indent n=n+1; \\ 
\hspace{1mm}\indent elseif pn$>$.8 \& (size(ttImg,1)*size(ttImg,2))$<$340 \textcolor{green}{\% probably part of a vehicle }\\ 
\hspace{1mm}\indent \indent \indent disp(\textcolor{red}{'Junk Filter'}); \\ 
\hspace{1mm}\indent \indent \indent \textcolor{blue}{if} n$>$2 \\ 
\hspace{1mm}\indent \indent \indent \indent bimg=ikmeans(bimg$>$0,n-1); \\ 
\hspace{1mm}\indent \indent \indent \textcolor{blue}{end} \\ 
\hspace{1mm}\indent \indent \indent xmin=[]; \\ 
\hspace{1mm}\indent \indent \indent xmax=[]; \\ 
\hspace{1mm}\indent \indent \indent ymin=[]; \\ 
\hspace{1mm}\indent \indent \indent ymax=[]; \\ 
\hspace{1mm}\indent \indent \indent k=1; \\ 
\hspace{1mm}\indent \indent \indent n=n-1; \\ 
\hspace{1mm}\indent \indent \indent  \\ 
\hspace{1mm}\indent elseif (size(ttImg,1)*size(ttImg,2))$<$250 \\ 
\hspace{1mm}\indent \indent \indent bimg=bimg-k*(bimg==k); \textcolor{green}{\%remove section }\\ 
\hspace{1mm}\indent \indent \indent k=k+1; \\ 
\hspace{1mm}\indent \textcolor{blue}{else} \\ 
\hspace{1mm}\indent \indent \indent  \\ 
\hspace{1mm}\indent \indent \indent xmin(end+1)=min(xI); \\ 
\hspace{1mm}\indent \indent \indent xmax(end+1)=max(xI); \\ 
\hspace{1mm}\indent \indent \indent ymin(end+1)=min(yI); \\ 
\hspace{1mm}\indent \indent \indent ymax(end+1)=max(yI); \\ 
\hspace{1mm}\indent \indent \indent \textcolor{blue}{if} dp \\ 
\hspace{1mm}\indent \indent \indent \indent line([xmin(end) xmax(end)],[ymin(end) ymin(end)]); \\ 
\hspace{1mm}\indent \indent \indent \indent line([xmin(end) xmax(end)],[ymax(end) ymax(end)]); \\ 
\hspace{1mm}\indent \indent \indent \indent line([xmin(end) xmin(end)],[ymin(end) ymax(end)]); \\ 
\hspace{1mm}\indent \indent \indent \indent line([xmax(end) xmax(end)],[ymin(end) ymax(end)]); \\ 
\hspace{1mm}\indent \indent \indent \textcolor{blue}{end} \\ 
\hspace{1mm}\indent \indent \indent k=k+1; \\ 
\hspace{1mm}\indent \textcolor{blue}{end} \\ 
\hspace{1mm}\textcolor{blue}{end} \\ 
}
\subsection{Identification Code}
\texttt{\hspace{1mm}\textcolor{blue}{function} identifycars(imgs,realim,obdb) \\ 
\hspace{1mm}\textcolor{green}{\%identifycars(imgs,realim,obdb) }\\ 
\hspace{1mm}colormap(\textcolor{red}{'bone'}); \\ 
\hspace{1mm}\textcolor{blue}{for} k=1:size(imgs,3) \\ 
\hspace{1mm}\indent [xmin,xmax,ymin,ymax]=findcars(imgs(:,:,k),0); \\ 
\hspace{1mm}\indent imshow(realim(:,:,:,k)); \\ 
\hspace{1mm}\indent \textcolor{blue}{for} j=1:length(xmin) \\ 
\hspace{1mm}\indent \indent \indent  \\ 
\hspace{1mm}\indent \indent \indent line([xmin(j) xmax(j)],[ymin(j) ymin(j)]); \\ 
\hspace{1mm}\indent \indent \indent line([xmin(j) xmax(j)],[ymax(j) ymax(j)]); \\ 
\hspace{1mm}\indent \indent \indent line([xmin(j) xmin(j)],[ymin(j) ymax(j)]); \\ 
\hspace{1mm}\indent \indent \indent line([xmax(j) xmax(j)],[ymin(j) ymax(j)]); \\ 
\hspace{1mm}\indent \indent \indent \textcolor{green}{\%[xmin(j),xmax(j),ymin(j),ymax(j)] }\\ 
\hspace{1mm}\indent \indent \indent \textcolor{green}{\%figure }\\ 
\hspace{1mm}\indent \indent \indent \textcolor{green}{\%imagesc(imgs(ymin(j):ymax(j),xmin(j):xmax(j),k)) }\\ 
\hspace{1mm}\indent \indent \indent cImg=imgs(ymin(j):ymax(j),xmin(j):xmax(j),k); \\ 
\hspace{1mm}\indent \indent \indent fi=featureim(cImg); \\ 
\hspace{1mm}\indent \indent \indent c=cov(reshape(fi,size(fi,1)*size(fi,2),size(fi,3))); \\ 
\hspace{1mm}\indent \indent \indent dists=calcdist(obdb,c); \\ 
\hspace{1mm}\indent \indent \indent obj=find(dists==min(dists)); \\ 
\hspace{1mm}\indent \indent \indent obj=obj(1); \\ 
\hspace{1mm}\indent \indent \indent \textcolor{blue}{switch} obj \\ 
\hspace{1mm}\indent \indent \indent \indent \textcolor{blue}{case} 1 \\ 
\hspace{1mm}\indent \indent \indent \indent \indent txt=\textcolor{red}{'Car'}; \\ 
\hspace{1mm}\indent \indent \indent \indent \textcolor{blue}{case} 2 \\ 
\hspace{1mm}\indent \indent \indent \indent \indent txt=\textcolor{red}{'Truck'}; \\ 
\hspace{1mm}\indent \indent \indent \indent \textcolor{blue}{case} 3 \\ 
\hspace{1mm}\indent \indent \indent \indent \indent txt=\textcolor{red}{'Multiple'}; \\ 
\hspace{1mm}\indent \indent \indent \indent \textcolor{blue}{case} 4 \\ 
\hspace{1mm}\indent \indent \indent \indent \indent txt=\textcolor{red}{'Junk'}; \\ 
\hspace{1mm}\indent \indent \indent \indent \textcolor{blue}{otherwise} \\ 
\hspace{1mm}\indent \indent \indent \indent \indent txt=\textcolor{red}{'err'}; \\ 
\hspace{1mm}\indent \indent \indent \textcolor{blue}{end} \\ 
\hspace{1mm}\indent \indent \indent text((xmin(j)+xmax(j))/2,(ymin(j)+ymax(j))/2,txt,\textcolor{red}{'Color'},[1 0 0]) \\ 
\hspace{1mm}\indent \indent \indent  \\ 
\hspace{1mm}\indent \textcolor{blue}{end} \\ 
\hspace{1mm}\indent menu(\textcolor{red}{'Ready?'},\textcolor{red}{'Yes'}); \\ 
\hspace{1mm}\textcolor{blue}{end} \\ 
}
\subsection{K-means Segmentation Code}
\texttt{\hspace{1mm}\textcolor{blue}{function} [segimage]=ikmeans(img,groups) \\ 
\hspace{1mm}[x,y]=find(img); \\ 
\hspace{1mm}u=kmeans([x';y']',groups,\textcolor{red}{'dist'},\textcolor{red}{'city'}); \\ 
\hspace{1mm}segimage=zeros(size(img,1),size(img,2)); \\ 
\hspace{1mm}\textcolor{blue}{for} k=1:groups \\ 
\hspace{1mm}\indent \textcolor{green}{\%segimage(x(u==k),y(u==k))=k; }\\ 
\hspace{1mm}\indent xk=x(u==k); \\ 
\hspace{1mm}\indent yk=y(u==k); \\ 
\hspace{1mm}\indent \textcolor{blue}{for} z=1:length(xk) \\ 
\hspace{1mm}\indent \indent \indent segimage(xk(z),yk(z))=k; \\ 
\hspace{1mm}\indent \textcolor{blue}{end} \\ 
\hspace{1mm}\textcolor{blue}{end} \\ 
}
\subsection{Feature Vector/Image Creation Code}
\texttt{\hspace{1mm}\textcolor{blue}{function} fd=featureim(k); \\ 
\hspace{1mm}mtlabd=0; \\ 
\hspace{1mm}fd=[]; \\ 
\hspace{1mm}\textcolor{green}{\% position values }\\ 
\hspace{1mm}xax=[1:size(k,1)]'*ones(1,size(k,2)); \\ 
\hspace{1mm}yax=([1:size(k,2)]'*ones(1,size(k,1)) )'; \\ 
\hspace{1mm}\indent sobel=[[-1 2 -1];[0 0 0];[1 2 1]]; \\ 
\hspace{1mm}\indent lap1=[[0 -1 0];[-1 4 -1];[0 -1 0]]; \\ 
\hspace{1mm}\indent lap2=[[-1 -1 -1];[-1 8 -1];[-1 -1 -1]]; \\ 
\hspace{1mm}\textcolor{green}{\%fd(:,:,1)=sqrt(xax.\texttt{\^}2+yax.\texttt{\^}2); }\textcolor{green}{\% First axis is distance from center }\\ 
\hspace{1mm}fd(:,:,end)=xax; \\ 
\hspace{1mm}fd(:,:,end+1)=yax; \\ 
\hspace{1mm}\textcolor{green}{\%image intesity }\\ 
\hspace{1mm}fd(:,:,end+1)=k; \\ 
\hspace{1mm} \\ 
\hspace{1mm}\textcolor{blue}{if} mtlabd==1 \textcolor{green}{\% diff command }\\ 
\hspace{1mm}\indent \textcolor{green}{\% first order derivative }\\ 
\hspace{1mm}\indent fd(:,:,end+1)=zeros(size(k,1),size(k,2)); \\ 
\hspace{1mm}\indent fd(2:end,:,end)=abs(diff(k,1,1)); \\ 
\hspace{1mm}\indent fd(:,:,end+1)=zeros(size(k,1),size(k,2)); \\ 
\hspace{1mm}\indent fd(:,2:end,end)=abs(diff(k,1,2)); \\ 
\hspace{1mm}\indent \textcolor{green}{\% second order derivatives }\\ 
\hspace{1mm}\indent fd(:,:,end+1)=zeros(size(k,1),size(k,2)); \\ 
\hspace{1mm}\indent fd(3:end,:,end)=abs(diff(k,2,1)); \\ 
\hspace{1mm}\indent fd(:,:,end+1)=zeros(size(k,1),size(k,2)); \\ 
\hspace{1mm}\indent fd(:,3:end,end)=abs(diff(k,2,2)); \\ 
\hspace{1mm}elseif mtlabd==2 \\ 
\hspace{1mm} \\ 
\hspace{1mm}\indent fd(:,:,end+1)=conv2(k,sobel,\textcolor{red}{'same'}); \textcolor{green}{\%vertical }\\ 
\hspace{1mm}\indent fd(:,:,end+1)=conv2(k,sobel',\textcolor{red}{'same'}); \textcolor{green}{\%horizontal }\\ 
\hspace{1mm}\indent \textcolor{green}{\% fd(:,:,end+1)=conv2(k,lap1,'same'); }\textcolor{green}{\%first laplacian approximation }\\ 
\hspace{1mm}\indent fd(:,:,end+1)=conv2(k,lap2,\textcolor{red}{'same'}); \textcolor{green}{\%second laplacian approximation (includes diagonals) }\\ 
\hspace{1mm}\textcolor{blue}{else} \\ 
\hspace{1mm}\indent fd(:,:,end+1)=conv2(k,lap2,\textcolor{red}{'same'}); \\ 
\hspace{1mm}\indent fd(:,:,end+1)=edge(k,\textcolor{red}{'canny'},.2); \\ 
\hspace{1mm}\textcolor{blue}{end} \\ 
}
\subsection{Covariance Matrix from Feature Vector Code}
\texttt{\hspace{1mm}\textcolor{blue}{function} c=gencov(img) \\ 
\hspace{1mm}figure(1) \\ 
\hspace{1mm}imagesc(img); \\ 
\hspace{1mm}[y,x]=ginput(2); \\ 
\hspace{1mm}x=round(x); \\ 
\hspace{1mm}y=round(y); \\ 
\hspace{1mm}fi=featureim(img(x(1):x(2),y(1):y(2))); \\ 
\hspace{1mm}c=cov(reshape(fi,size(fi,1)*size(fi,2),size(fi,3))); \\ 
}
\end{document}